\documentclass{article} %
\usepackage[final]{colm2026_conference}

\usepackage{microtype}
\usepackage{hyperref}
\usepackage{url}
\usepackage{booktabs}

\newcommand{\method}{Jagle}

\newcommand{\resourcelinks}{%
  \par\smallskip
  {\normalsize
  \begin{tabular}{@{}rl@{}}
    \raisebox{-0.15em}{\includegraphics[height=1.1em]{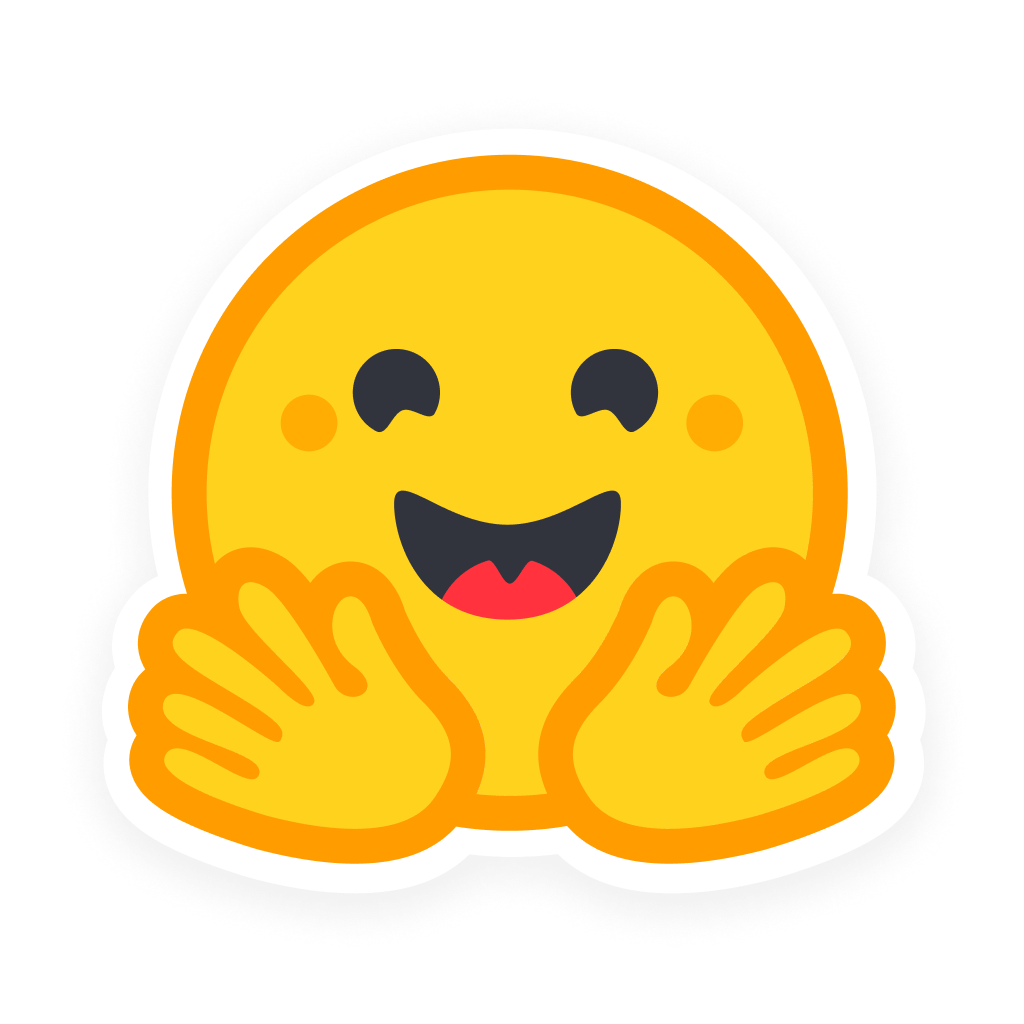}} &
    \textbf{Dataset: }
    \href{https://huggingface.co/datasets/llm-jp/Jagle}{llm-jp/Jagle}
    \\[0.3em]
    \raisebox{-0.15em}{\includegraphics[height=1.1em]{assets/hf-logo.png}} &
    \textbf{Models: }
    \href{https://huggingface.co/collections/llm-jp/jagle}{Jagle-VL-2.2B Collection} ~$\cdot$~ \href{https://huggingface.co/llm-jp/llm-jp-4-vl-9b-beta}{llm-jp/llm-jp-4-vl-9b-beta}
    \\[0.3em]
    \raisebox{-0.15em}{\includegraphics[height=1.1em]{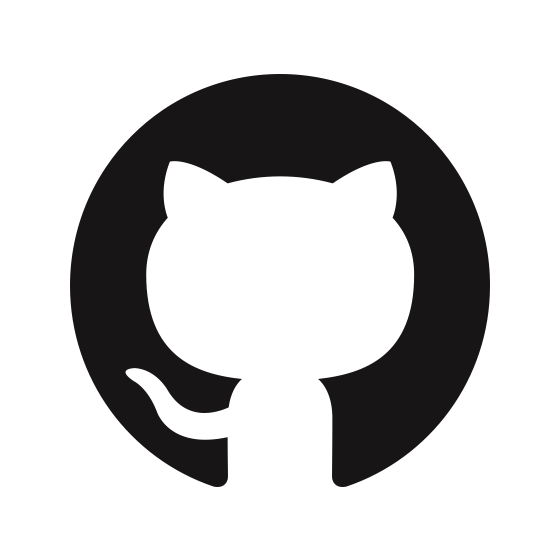}} &
    \textbf{Code: }
    \href{https://github.com/llm-jp/llm-jp-4-vl}{GitHub Repository}
  \end{tabular}
  }
  \par\smallskip
}

\usepackage{multirow}
\usepackage{multicol}

\usepackage{tcolorbox}
\tcbuselibrary{breakable, skins}
\usepackage{tikz}
\usepackage{comment} %

\usepackage{amssymb}
\usepackage{xcolor}
\usepackage{colortbl}
\usepackage[whole]{bxcjkjatype}

\newtcolorbox[auto counter, number within=section]{Prompt}[2][]{%
  colback=white, %
  width=\linewidth, %
  arc=3mm,
  boxrule=0.8mm, %
  title=\large #2, %
  breakable, %
  fonttitle=\small, %
  fontupper=\footnotesize, %
  #1 %
}

\newtcolorbox{SlimCode}{
  enhanced,
  colback=gray!20,
  colframe=gray!50,
  boxrule=0.4pt,
  arc=2mm,
  outer arc=2mm,
  left=2mm,
  right=2mm,
  top=1mm,
  bottom=1mm,
  boxsep=2pt,
  breakable,
  listing only,
  listing options={
    basicstyle=\ttfamily\scriptsize,  %
    breaklines=true,
    columns=fullflexible,
  },
  before skip=5pt,   %
  after skip=5pt     %
}

\usepackage{wrapfig}

\AtBeginEnvironment{SlimCode}{\footnotesize}

\usepackage{lineno}

\definecolor{darkblue}{rgb}{0, 0, 0.5}
\hypersetup{colorlinks=true, citecolor=darkblue, linkcolor=darkblue, urlcolor=darkblue}

\title{Jagle: Building a Large-Scale Japanese Multimodal \\Post-Training Dataset for Vision-Language Models}

\author{Issa Sugiura$^{1,2}$ \quad
  Keito Sasagawa$^{3,2}$ \quad
  Keisuke Nakao$^{3,2}$ \quad
  Koki Maeda$^{1,2}$ \\
  {\bf Ziqi Yin$^{2}$ \quad
  Zhishen Yang$^{2}$ \quad
  Shuhei Kurita$^{4,2}$ \quad
  Yusuke Oda$^{2}$} \\
  {\bf Ryoko Tokuhisa$^{5,6}$ \quad
  Daisuke Kawahara$^{3,2}$ \quad
  Naoaki Okazaki$^{1,7,2}$} \\[2mm]
  $^1$Institute of Science Tokyo \quad
  $^2$NII LLMC \quad
  $^3$Waseda University \quad
  $^4$NII \\
  $^5$Aichi Institute of Technology \quad
  $^6$RIKEN \quad
  $^7$AIST
}

\begin{document}

\ifcolmsubmission
\linenumbers
\fi

\maketitle

\begin{abstract}
Developing vision-language models (VLMs) that generalize across diverse tasks requires large-scale training datasets with diverse content. In English, such datasets are typically constructed by aggregating and curating numerous existing visual question answering (VQA) resources. However, this strategy does not readily extend to other languages, where VQA datasets remain limited in both scale and domain coverage.
In this work, we introduce \textbf{Jagle}, the largest Japanese multimodal post-training dataset to date, comprising approximately 9.2 million instances across diverse tasks. Rather than relying on existing VQA datasets, we collect heterogeneous source data, including images, image-text pairs, and PDF documents, and generate VQA pairs through multiple strategies such as VLM-based QA generation, translation, and text rendering.
Experiments demonstrate that a 2.2B model trained with Jagle surpasses InternVL3.5-2B in average score across ten Japanese evaluation tasks and comes within five points of Qwen3-VL-2B-Instruct. Moreover, combining Jagle with FineVision improves English performance over training with FineVision alone.
These benefits generalize across model backbones and scales: LLM-jp-4-VL 9B beta, a 9B-scale model built on the Japanese-centric LLM-jp-4-8B-instruct, achieves Japanese performance comparable to Qwen3-VL-8B-Instruct.
To facilitate reproducibility and future research, we release the dataset, models, and code.
\resourcelinks
\end{abstract}

\section{Introduction}
Vision-Language Models (VLMs), which extend large language models (LLMs) with visual understanding capabilities, have recently achieved rapid progress~\citep{liu2023llava,openai2024gpt4ocard,bai2025qwen3vl}.
Both proprietary models~\citep{openai2025gpt5.1,google2025gemini3pro} and open-weight models~\citep{zhu2025internvl3,bai2025qwen3vl,kimiteam2026kimik25visualagentic} have demonstrated strong multimodal reasoning abilities across a wide range of tasks.

A key factor driving these advances is the availability of large-scale, high-quality training datasets~\citep{wiedmann2025finevision,li2025eagle2,tong2024cambrian}.
In the research community, the development of large-scale English multimodal post-training datasets has been particularly active~\citep{tong2024cambrian,nvidia2025nvidianemotronnanov2}.

A common approach for constructing such large-scale datasets is to collect, curate, balance, and unify the format of a large number of existing VQA datasets~\citep{tong2024cambrian,li2025eagle2}. For example, FineVision~\citep{wiedmann2025finevision} built a dataset of approximately 24 million instances by aggregating over 100 existing English datasets.
However, this approach is difficult to apply to other languages, where existing VQA datasets are far less abundant.

To address this limitation, we propose an alternative pipeline that collects diverse source data, including images, image-text pairs, and PDFs, and generates VQA pairs through a combination of strategies such as VLM-based QA generation, translation, and text rendering.
Focusing specifically on Japanese, we construct \textbf{Jagle}\footnote{The name Jagle is a blend of \textbf{Ja}panese and Ea\textbf{gle}2~\citep{li2025eagle2}.}, a large-scale Japanese multimodal post-training dataset\footnote{Following Eagle2~\citep{li2025eagle2}, we use the term ``post-training'' to refer to multimodal training performed after initializing from pretrained LLMs and vision encoders.} consisting of approximately 9.2 million instances spanning 5 task categories and 17 subsets.
Unlike prior approaches that rely on curating and aggregating existing datasets, our pipeline builds the dataset from scratch.
This design makes our methodology readily transferable to other low-resource languages where large-scale multimodal resources are limited, as is the case for most non-English languages.

As shown in Table~\ref{tab:dataset_comparison}, \method{} substantially expands both the scale and task diversity compared to existing Japanese multimodal post-training datasets, such as DEJIMA~\citep{katsube2025dejima} and LLM-jp-3 VILA~\citep{sasagawa-etal-2025-constructing}.

Through experiments, we verify the effectiveness of Jagle for improving the Japanese performance of VLMs.
We show that a 2.2B model trained on Jagle outperforms InternVL3.5-2B on the average score across 10 Japanese benchmarks and comes within 5 points of Qwen3-VL-2B-Instruct.
We further analyze the effect of mixing Jagle with the English dataset FineVision, and find that the average score across 10 English benchmarks exceeds that of a FineVision-only baseline, indicating that adding Jagle does not degrade but rather improves English task performance.
Moreover, the effectiveness of Jagle generalizes across model backbones and scales: LLM-jp-4-VL 9B beta, a 9B-scale model trained with the Japanese-centric LLM-jp-4-8B-instruct backbone, achieves Japanese performance comparable to Qwen3-VL-8B-Instruct.

To facilitate reproducibility and foster future research in VLMs, we publicly release our dataset, models, and code.

\begin{table}[t]
\centering
\setlength{\tabcolsep}{4pt}
\begin{tabular}{lcrrr}
\toprule
\textbf{Dataset} &\textbf{Language} & \textbf{Categories}& \textbf{Subsets} & \textbf{Examples} \\%& \textbf{寛容ライセンス}\\
\midrule
Cambrian-7B~\citep{tong2024cambrian}&English&9&70 & 7.1M  \\
FineVision~\citep{wiedmann2025finevision}&English & 9& 185 & 24.2M   \\%&  \xmark \\
\midrule
DEJIMA~\citep{katsube2025dejima} & Japanese & 2 & 2 & 3.9M\\
LLM-jp-3 VILA~\citep{sasagawa-etal-2025-constructing}& Japanese&3 & 4 & 0.4M \\%& \xmark\\ %
\textbf{\method{} (Ours)} & Japanese& \textbf{5} & \textbf{17} & \textbf{9.2M}\\%&\cmark\\
\bottomrule
\end{tabular}
\caption{Comparison of open multimodal post-training datasets. Task categories follow the Eagle2 taxonomy~\citep{li2025eagle2}. \method{} is the largest Japanese multimodal post-training dataset.}
\label{tab:dataset_comparison}
\end{table}

\section{Related Work}

\noindent\textbf{Post-training Datasets for VLMs.}
To develop VLMs with broad knowledge and the ability to handle diverse tasks, the construction of large-scale multimodal post-training datasets has been actively pursued~\citep{tong2024cambrian,li2025llavaonevision,li2025eagle2,wiedmann2025finevision}.
Early VLM research primarily focused on image captioning, with models typically trained on relatively small-scale datasets with limited task coverage~\citep{liu2023llava}.
As VLM capabilities have advanced, research has expanded beyond captioning to more diverse tasks, including chart and document comprehension~\citep{masry2022chartqa,Mathew2021docvqa}, as well as computer-use tasks involving interaction with graphical user interfaces~\citep{xie2024osworld}.
Accordingly, post-training datasets have grown substantially in both task diversity and scale, enabling broader multimodal reasoning abilities~\citep{tong2024cambrian,li2025llavaonevision,Deitke2025molmo}.
In early VLM research, a prominent approach was to synthesize captioning data using strong models such as GPT-4o~\citep{liu2023llava,lin2024sharegpt4v,openai2024gpt4ocard}.
Another line of work leverages the coding capabilities of LLMs to generate scripts for producing chart images, enabling the creation of VQA data in the chart domain~\citep{yang-etal-2025-scaling}.
More recently, large-scale English dataset construction has shifted toward collecting, curating, balancing, and unifying the format of the many existing English VQA datasets that have accumulated over the years, yielding large-scale, diverse, and high-quality resources~\citep{tong2024cambrian,li2025llavaonevision,li2025eagle2,wiedmann2025finevision}.

\noindent\textbf{Japanese multimodal post-training datasets.}
The development of Japanese post-training datasets for VLMs remains limited, and existing datasets are typically small in scale and lack sufficient task coverage.
For example, LLM-jp-3 VILA~\citep{sasagawa-etal-2025-constructing} does not cover practically important domains such as document understanding and chart comprehension.
As a result, the model trained on the dataset exhibits relatively weak performance on document and chart understanding benchmarks~\citep{maeda2026llm-jp-eval-mm}.
Furthermore, the dataset construction pipeline relies on proprietary models such as GPT-4o~\citep{openai2024gpt4ocard}, which introduces licensing constraints that limit practical usage.

\begin{figure}[t]
\begin{center}
\includegraphics[width=\linewidth]{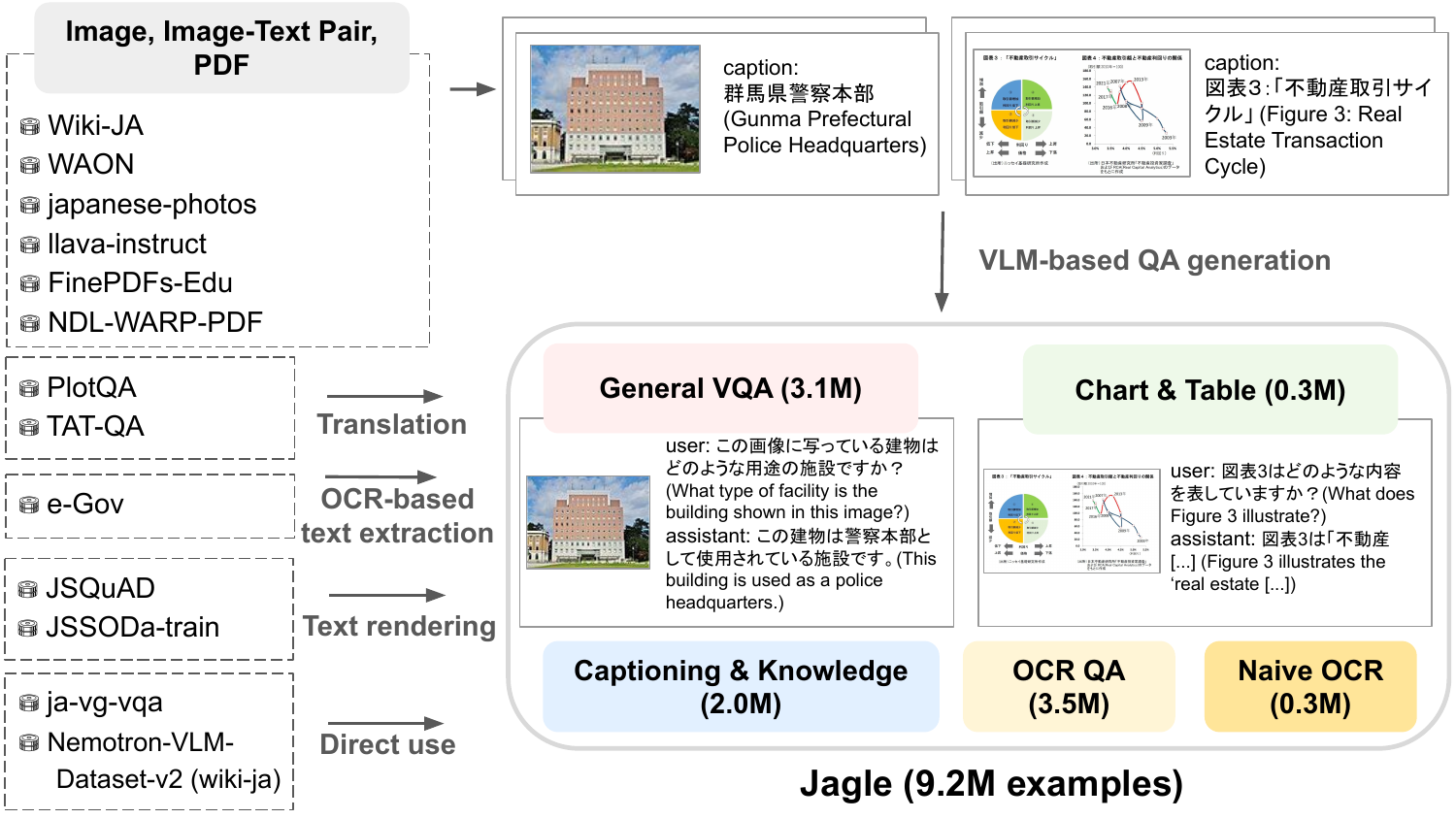}
\caption{Construction pipeline of Jagle. Our pipeline leverages diverse data sources, including images, image-text pairs, and PDF corpora, and integrates multiple QA generation strategies such as VLM-based QA generation, translation, OCR-based text extraction, text rendering, and direct utilization of existing data to produce VQA samples.}
\label{fig:construction-pipeline}
\end{center}
\end{figure}

\section{Construction of \method{}}

The construction pipeline of Jagle is shown in Figure~\ref{fig:construction-pipeline}. We build Jagle through three steps: (1) category definition, (2) source data collection, and (3) QA generation.

\subsection{Category definition}

To construct a post-training dataset that covers diverse tasks, we define five categories based on the taxonomy used in Eagle2~\citep{li2025eagle2}. Eagle2 defines nine categories: General VQA, Chart \& Table, Captioning \& Knowledge, OCR QA, Naive OCR, Grounding \& Counting, Math, Science, and Text-only.
From these, we select the following five categories as our target: General VQA, Chart \& Table, Captioning \& Knowledge, OCR QA, and Naive OCR. We exclude categories such as Math and Science, which are relatively less language-dependent.
The remaining categories are left for future exploration.

\subsection{Source data collection}

To construct a dataset that comprehensively covers the five target categories, it is necessary to collect appropriate source data for each category. For example, OCR QA benefits from text-rich sources such as PDF documents or natural images containing text, whereas General VQA relies on diverse natural images obtained from web-based resources. Below, we describe the data sources used for each category.

\noindent\textbf{General VQA.}
General VQA involves answering questions about images from diverse domains (e.g., people, objects, and scenes). To cover a wide range of visual content, we utilize six data sources. These include japanese-photos~\citep{ThePioneer2024japanese-photos}, a small-scale image corpus consisting of photos taken in Japan; Wiki-JA-Pair, a 1M image--text pair dataset we construct from Japanese Wikipedia articles~\citep{wikipedia_ja} and release as part of this work; and WAON~\citep{sugiura2025waon}, a 155M Japanese image-text pair dataset collected from Common Crawl.

\noindent\textbf{Chart \& Table.}
Chart \& Table is a task that requires extracting and interpreting information from charts and tables, including reading values and performing calculations when necessary. Constructing this category requires a large collection of such visual data. We utilize existing English datasets such as PlotQA and TAT-QA, as well as WAON.

To extract chart and table images from WAON, we retrieve image-text pairs whose captions begin with keywords such as ``図'' (chart) or ``表'' (table). The effectiveness of this heuristic is verified through manual inspection of the retrieved samples.

\noindent\textbf{Captioning \& Knowledge.}
This category involves generating descriptions of given images and capturing associated knowledge.
We collect both natural and document images from Wiki-JA-Pair and PDF files crawled from URLs provided by the National Diet Library's Web Archiving Project (NDL WARP)~\citep{ndl-warp-pdf}.

\noindent\textbf{OCR QA.}
OCR QA is a question-answering task focused on textual information contained within images. Constructing a dataset for this category requires images that are rich in text content. For this purpose, we use the Japanese subsets of PDF corpora such as NDL WARP and FinePDFs-Edu~\citep{kydlicek2025finepdfs}. Additionally, to cover text present in natural images, we also utilize the WAON dataset.

\noindent\textbf{Naive OCR.}
Naive OCR is a task that involves directly extracting text from images in reading order. For this purpose, we use PDFs from Japanese government agencies, collected via the e-Gov portal site, and convert them into images~\citep{egovjp}.
In addition, we incorporate the Wiki-JA subset of Nemotron-VLM-Dataset-v2, which constructs OCR tasks by rendering Japanese Wikipedia articles as images~\citep{nvidia2025nvidianemotronnanov2}.

\subsection{QA generation}

We construct QA pairs using several approaches: (1) VLM-based QA generation, (2) OCR-based text extraction, (3) text rendering, and (4) translation. To ensure the quality of the generated data, we manually inspect and analyze a subset of the synthesized VQA pairs and iteratively refine the generation process.
To enhance dataset diversity, we minimize the inclusion of similar images. Specifically, when using PDF data, we randomly select only one page per PDF. Additionally, for data sources containing visually similar images, we perform deduplication before generating QA pairs.

\noindent\textbf{VLM-based QA generation.}
While human annotation is ideal for constructing VQA datasets, scaling it to the volume required for VLM training is impractical.
Traditional approaches rely on template-based generation~\citep{masry2022chartqa}; however, such methods tend to produce rigid question formats, which can lead to overfitting to specific patterns.
Recent studies have demonstrated that leveraging existing VLMs to generate QA pairs enables large-scale synthesis without constraining question formats~\citep{lin2024sharegpt4v,liu2023llava}.
Specifically, we use Qwen3-VL, a high-performing open-weight model with a permissive license and strong performance in both Japanese and English~\citep{sugiura2026jammeval}. In most cases, we employ Qwen3-VL-235B-A22B-Instruct, the strongest instruct model in the Qwen3-VL series.
To generate QA pairs using the model, we design task-specific instruction prompts for each dataset, and iteratively refine them through manual inspection of the generated QA pairs.\footnote{Detailed prompts used for QA generation are provided in Appendix~\ref{sec:prompt_for_qa_generation}.}
When available (e.g., in image-text pairs), captions are included in the prompt to provide knowledge not contained in the VLM used for QA generation, such as associations between faces and person names, enabling the construction of a dataset that does not overly rely on the model's implicit knowledge.
We leverage vLLM~\citep{woosuk2023vLLM} for inference, enabling efficient generation of QA pairs.

\noindent\textbf{OCR-based text extraction.}
For OCR-related tasks that require accurate extraction of text from images, obtaining precise textual information is essential. We employ PaddleOCR-VL~\citep{cui2025paddleocrvl}, an OCR-specialized model, to extract text and generate QA pairs.

\noindent\textbf{Text rendering.}
Text rendering, which involves converting prepared text into images and automatically generating QA pairs using templates, is a common approach for constructing Naive OCR VQA datasets~\citep{nvidia2025nvidianemotronnanov2}. This method allows precise control over the textual content. We generate VQA instances by rendering text into JPG images using text-based QA datasets such as JSQuAD~\citep{kurihara-etal-2022-jglue}.
JSSODa~\citep{sasagawa-etal-2026-evaluating}, a synthetic OCR dataset, is constructed in a similar manner.

\noindent\textbf{Translation.}
Translating English datasets into other languages is a common approach for leveraging large-scale English resources in multilingual settings~\citep{sugiura2025waon,penedo2026finetranslations}. However, when translating multimodal data, special care is required to maintain consistency between the language of the text in the image and that of the QA pairs. In this work, we construct VQA datasets by translating English datasets such as Plot-QA~\citep{Methani_2020_plotqa} and TAT-QA~\citep{zhu-etal-2021-tatqa}, which generate chart and table images via Python scripts, using LLM-jp-3-13B-instruct~\citep{llmjp2024llmjp}, an LLM trained on large-scale Japanese and English corpora.
By translating the English text contained in the chart metadata, such as table cell values and axis labels, and then re-rendering the charts and tables using the Plotly library, we generate Japanese versions of the original images that remain consistent with the corresponding QA pairs.
Because the images are re-rendered directly from the translated metadata rather than from translated plotting scripts, and because the translated content consists of relatively short and structured text (e.g., questions, answers, and chart labels), translation errors are easy to control; manual inspection of the translated subsets revealed no rendering failures or Japanese fluency issues.

\section{Exploring Jagle}

In this section, we analyze the Jagle dataset through statistics and qualitative case studies. An additional analysis of image embeddings is provided in Appendix~\ref{sec:image_embedding_analysis}.

\begin{table}[t]
\centering
\small
\setlength{\tabcolsep}{2pt}
\begin{tabular}{llrp{4cm}l}
\toprule
\textbf{Category} &\textbf{Subset Name}  & \textbf{Samples} & \textbf{Data Source}&\textbf{Method}\\
\midrule
\multirow{6}{*}{General VQA}&
 japanese-photos-VQA & 1,163 &  japanese-photos~{\tiny\citep{ThePioneer2024japanese-photos}} & VLM-based\\
& JSQuAD-Vision-filterd-15k & 14,790 &JSQuAD~{\tiny\citep{kurihara-etal-2022-jglue}}&Text rendering\\
& ja-vg-vqa & 99,202 &   ja-vg-vqa~{\tiny\citep{shimizu-etal-2018-visual}}&Direct use\\
& llava-instruct-ja-qwen3vl & 155,657 &llava-instruct~{\tiny\citep{liu2023llava}}&VLM-based\\
& WIKI-JA-VQA & 936,871 &  Wiki-JA-Pair (ours)  & VLM-based\\
& WAON-VQA &1,912,226  & WAON~{\tiny\citep{sugiura2025waon}}&VLM-based\\
\midrule
\multirow{3}{*}{Chart \& Table}
& tat-qa-ja-translated-2k & 2,180 & TAT-QA~{\tiny\citep{zhu-etal-2021-tatqa}} & Translation\\
&WAON-Chart-VQA  & 98,791 &  WAON~{\tiny\citep{sugiura2025waon}}&VLM-based\\
& plotqa-ja-translated-153k & 152,912 & Plot-QA~{\tiny\citep{Methani_2020_plotqa}}&Translation\\
\midrule
Captioning & WIKI-JA-Captioning  & 993,006& Wiki-JA-Pair (ours)&VLM-based\\
\& Knowledge & NDL-PDF-detail  & 1,000,000 &  NDL WARP PDF~{\tiny\citep{ndl-warp-pdf}}&VLM-based\\
\midrule
\multirow{3}{*}{OCR QA} &WAON-OCR-VQA  & 871,847 & WAON~{\tiny\citep{sugiura2025waon}}&VLM-based\\
&NDL-PDF-simple & 1,000,000 &  NDL WARP PDF~{\tiny\citep{ndl-warp-pdf}}&VLM-based\\
&FinePDFs-Edu-JA-VQA& 1,666,699 &  FinePDFs-Edu~{\tiny\citep{kydlicek2025finepdfs}}&VLM-based\\
\midrule
 \multirow{4}{*}{Naive OCR} &JSSODa-train-18k-v2 & 17,991 &  JSSODa~{\tiny\citep{sasagawa-etal-2026-evaluating}}&Text rendering\\
&e-Gov-OCR  & 31,190 &  e-Gov~{\tiny\citep{egovjp}}&OCR-based\\
&Nemotron-Wiki-Ja-OCR & 199,999 & Nemotron-VLM-Dataset-v2 (wiki-ja)~{\tiny\citep{nvidia2025nvidianemotronnanov2}} & Direct use\\
\midrule
Total & & 9,154,524&  & \\
\bottomrule
\end{tabular}
\caption{Statistics of each subset in \method{}: samples, data source, and QA construction method.}
\label{tab:dataset}
\end{table}

\begin{table}[t]
\centering
\small
\begin{tabular}{lrrrr}
\toprule
\textbf{Dataset}&\textbf{Samples}&\textbf{Unique Images}&\textbf{Turns}&\textbf{Answer Tokens}\\
\midrule
FineVision~{\scriptsize\citep{wiedmann2025finevision}}& 24,210,611 & 15,606,828&88,356,316&10,146,129,067\\
Jagle (Ours) &9,154,524&9,051,232&58,906,972&3,510,910,782\\
\bottomrule
\end{tabular}
\caption{Comparison of dataset sizes between \method{} and FineVision~\citep{wiedmann2025finevision} in terms of samples, unique images, turns, and answer tokens.}
\label{tab:dataset-finevision-vs-jagle}
\end{table}

\begin{figure}[t]
\begin{center}
\includegraphics[width=\linewidth]{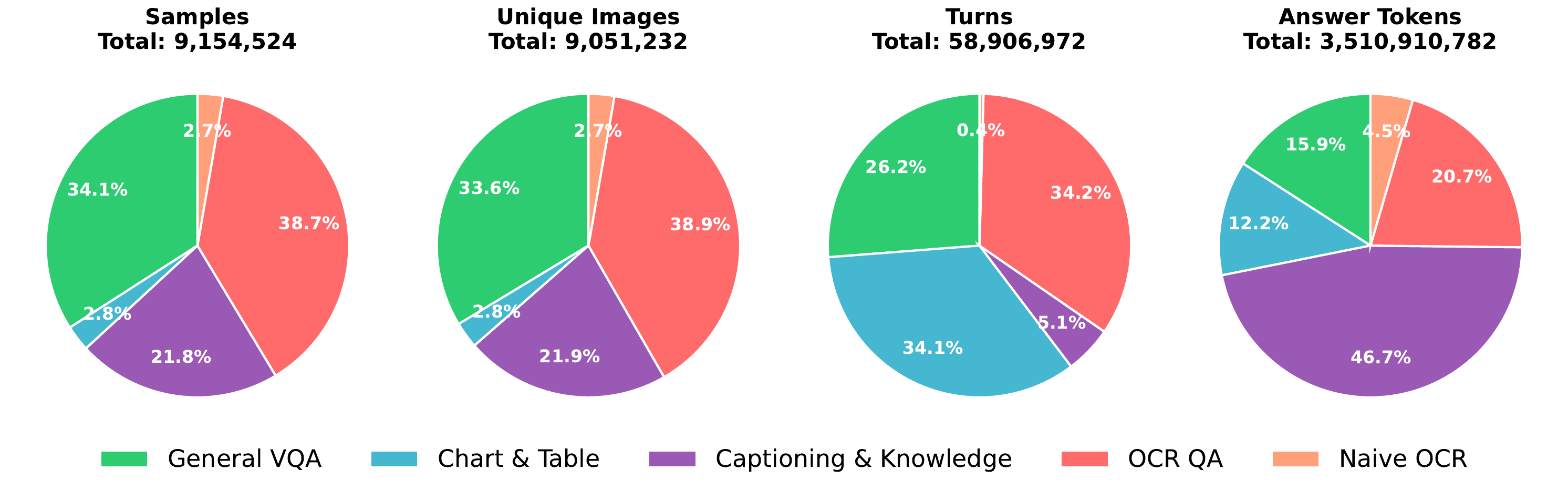}
\caption{Category distribution of \method{} across four metrics: number of samples, unique images, turns, and answer tokens.}
\label{fig:ja-dataset-piechart}
\end{center}
\end{figure}

\subsection{Statistics of \method{}}
Table~\ref{tab:dataset} presents the number of samples, data sources, and QA construction methods for each subset of Jagle.\footnote{Appendix~\ref{sec:detailed_statistic_jagle} provides detailed statistics for each subset, including the number of samples, unique images, turns, and answer tokens.} \method{} consists of approximately 9.2 million instances spanning five categories and 17 subsets.

Table~\ref{tab:dataset-finevision-vs-jagle} compares statistics such as the number of samples, unique images, turns, and answer tokens between FineVision and Jagle. Jagle is approximately 2.6 times smaller than FineVision, the largest English dataset in terms of scale; given that the English-to-Japanese ratio in Common Crawl is roughly 9:1, this suggests that Jagle is sufficiently large in scale.

\subsection{Category distribution}
Figure~\ref{fig:ja-dataset-piechart} shows the category distribution, weighted by the number of samples, answer tokens, turns, and images. We use the Qwen3~\citep{yang2025qwen3} tokenizer to compute the number of answer tokens.
In terms of sample count, General VQA, OCR QA, and Captioning \& Knowledge each account for more than 20\% of the dataset.
When weighted by the number of answer tokens, the Captioning \& Knowledge category occupies a large proportion, primarily due to detailed captioning data such as NDL-PDF-detail, which contains long-form descriptions.
For the distribution over the number of turns, Chart \& Table accounts for a large proportion. This is because datasets such as PlotQA and TAT-QA associate multiple question-answer pairs with a single image; aggregating them into a multi-turn format results in a higher number of turns per instance.

\begin{figure}[t]
\begin{center}
\includegraphics[width=\linewidth]{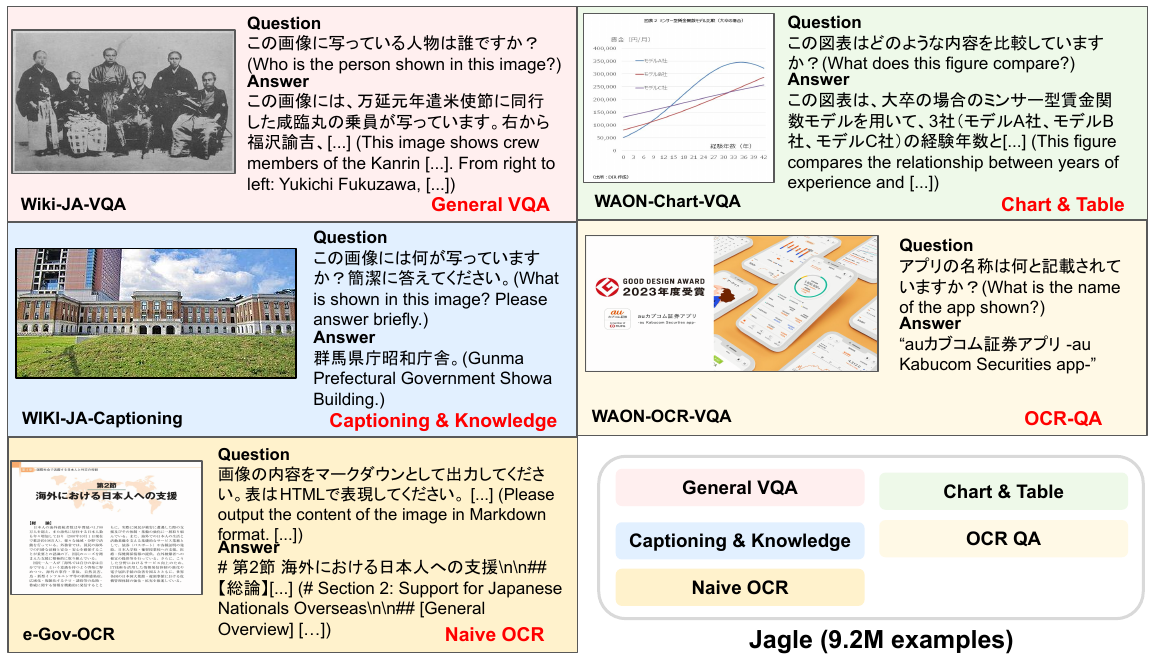}
\caption{Representative VQA examples from each category in \method{}. The dataset covers a wide variety of visual content, including natural images, charts and tables, document images, and presentation slides.}
\label{fig:jagle_cases}
\end{center}
\end{figure}

\subsection{Qualitative case studies}

As a qualitative examination of VQA examples in Jagle, we present representative examples randomly selected from each category in Figure~\ref{fig:jagle_cases}. Jagle includes a diverse set of VQA instances covering not only captioning and general question answering on natural images, but also question answering on charts and tables, document images, and slide-based visuals. Training on such diverse visual representations is expected to enable models to generalize across a wide range of image types.

\section{Experiments}

To evaluate the effectiveness of \method{}, we train a 2.2B VLM on three data settings: Jagle alone to directly measure its effect, FineVision alone as an English-only baseline, and their mixture to reflect the common practice of training recent VLMs on multilingual data.
\subsection{Training setup}
\label{sec:training_setup}

\noindent\textbf{Model Architecture.}
We use Qwen3-1.7B-Instruct~\citep{yang2025qwen3} as the LLM and SigLIP2-so400m-patch16-512~\citep{tschannen2025siglip2} as the image encoder.
For the multimodal projector, we employ a two-layer MLP. The resulting model contains approximately 2.2B parameters.
We refer to the models trained in this setup as Jagle-VL-2.2B, naming each model by appending its training data (e.g., Jagle-VL-2.2B-Jagle-FineVision).
We adopt the OpenAI Harmony format~\citep{openai2025gptoss} as the chat template.
To support high-resolution images, we employ dynamic tiling~\citep{li2025llavaonevision}, which splits an image into multiple tiles, extracts visual tokens from each tile, and concatenates them before feeding them into the LLM.

\noindent\textbf{Hyperparameters.}
We adopt a single-stage training strategy, following~\citet{wiedmann2025finevision}.
While some prior works such as VILA~\citep{Lin2024VILA} adopt multi-stage pipelines that first train on image--text pairs and then perform instruction tuning on VQA-style data, FineVision~\citep{wiedmann2025finevision} has shown that single-stage training directly on instruction-style multimodal data can also be effective.
We adopt the single-stage setup to isolate the effect of dataset quality and composition while avoiding confounding factors from multi-stage curricula and additional hyperparameter choices.
The LLM and vision encoder are initialized from pre-trained weights, while the multimodal projector is randomly initialized. We update all model parameters throughout training. We train for 60{,}000 steps with a batch size of 1{,}024 and a maximum sequence length of 4{,}096 tokens.
For the learning rate schedule, we adopt a Warmup--Stable--Decay scheme~\citep{wen2025understanding}. The peak learning rates are set to $2 \times 10^{-5}$ for the LLM and vision encoder and $1 \times 10^{-4}$ for the multimodal projector. The learning rates are linearly warmed up to their respective peak values over the first 2{,}000 steps, held constant during the stable phase, and then linearly decayed from 80\% of the total training steps to $0.1\times$ their peak values.
A full training run takes approximately 72 hours on 128 H200 GPUs. For the Jagle + FineVision setting, 60{,}000 training steps correspond to approximately two epochs.

\subsection{Evaluation setup}

\noindent\textbf{Evaluation Datasets.}
We evaluate models on a diverse set of English and Japanese benchmarks covering a wide range of tasks.
For English evaluation, we use 10 benchmarks: AI2D~\citep{kembhavi2016ai2d}, ChartQA~\citep{masry2022chartqa}, DocVQA~\citep{Mathew2021docvqa}, InfoVQA~\citep{Mathew2022infovqa}, OK-VQA~\citep{Marino2019okvqa}, RealWorldQA~\citep{xai2024realworldqa}, ScienceQA~\citep{lu2022scienceqa}, TextVQA~\citep{Singh2019textvqa}, BLINK~\citep{fu2024blink}, and MMMU~\citep{yue2023mmmu}.
For Japanese evaluation, we use 10 benchmarks: Heron-Bench~\citep{inoue2024heronbench}, JA-VLM-Bench-In-the-Wild~\citep{akiba2025evo}, JA-Multi-Image-VQA~\citep{inoue2024jamultiimage}, JGraphQA~\citep{jgraphqa}, CC-OCR-JA~\citep{yang2024ccocr}, CVQA-JA~\citep{mogrovejo2024cvqa}, JDocQA~\citep{onami2024jdocqa}, MECHA-JA~\citep{maeda2025mecha}, BusinessSlideVQA~\citep{stockmark2025businessslidevqa}, and JMMMU~\citep{onohara2025jmmmu}.
For the first seven Japanese benchmarks, we use the refined versions provided by JAMMEval~\citep{sugiura2026jammeval}, which correct issues such as ambiguity and incorrect answers in the original datasets.

\noindent\textbf{Baselines.}
We compare our models with two strong open-weight multilingual vision-language baselines of comparable scale: Qwen3-VL-2B-Instruct~\citep{bai2025qwen3vl} and InternVL3.5-2B~\citep{wang2025internvl35}.
We additionally compare with LLM-jp-3 VILA 14B~\citep{sasagawa-etal-2025-constructing}, a model trained on a prior Japanese dataset, and with 8B-scale VLMs (Qwen3-VL-8B-Instruct and InternVL3.5-8B) in Section~\ref{sec:generality_backbones}.

\noindent\textbf{Evaluation Protocol.}
For all evaluated models, we set the decoding temperature to 0, and the maximum number of generated tokens is set sufficiently large for each task.
For short answer format tasks, we use GPT-5.1 (\texttt{gpt-5.1-2025-11-13})~\citep{openai2025gpt5.1} as the judge model.
Each evaluation is run three times, and we report the mean.
Run-to-run variation arises only for tasks that use LLM-based scoring; as shown in Appendix~\ref{sec:per_task_results}, the standard deviations over three runs are substantially smaller than the performance gaps between models, indicating that our findings are robust to evaluation noise.

\begin{figure}[t]
\begin{center}
\includegraphics[width=\linewidth]{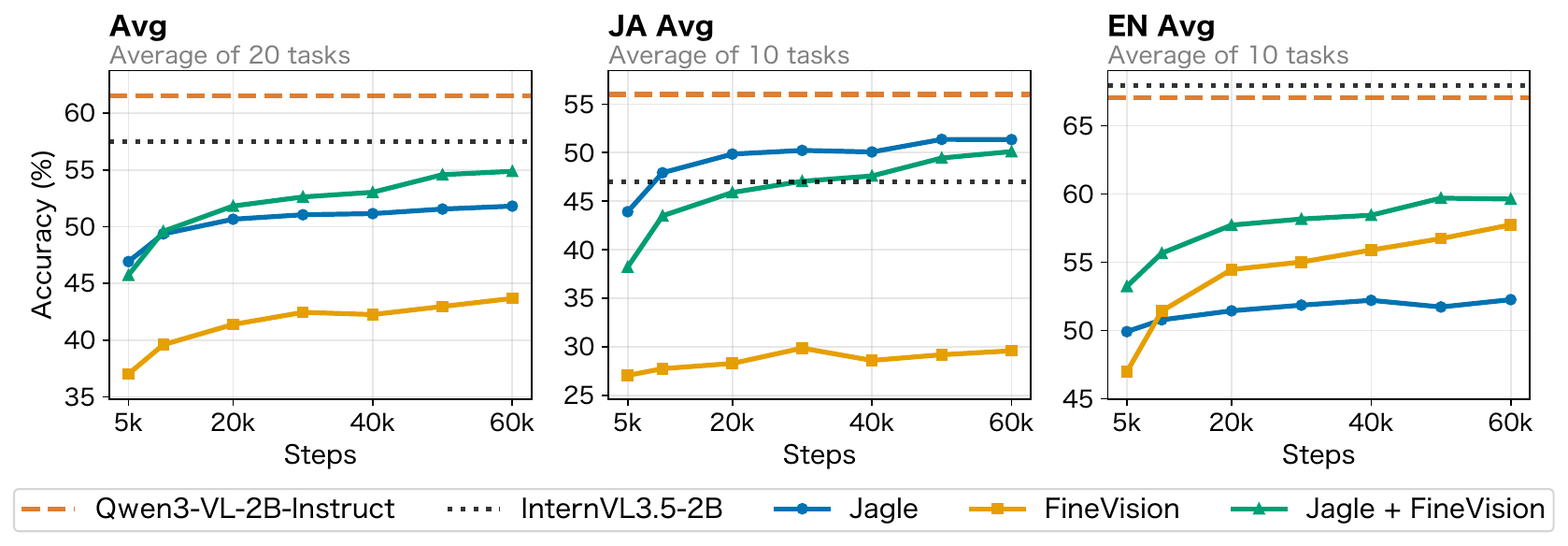}
\caption{Training dynamics under each data setting for the macro-averaged score over all 20 tasks (Avg), 10 Japanese tasks (JA Avg), and 10 English tasks (EN Avg). The model trained on \method{} outperforms the model trained on FineVision by over 20 points on JA Avg.}
\label{fig:training_curve_avg}
\end{center}
\end{figure}

\subsection{Results}
Figure~\ref{fig:training_curve_avg} shows training dynamics across dataset settings measured by the macro-averaged score over all 20 tasks (Avg), together with averages over the 10 Japanese tasks (JA Avg) and 10 English tasks (EN Avg).
Baseline scores of existing models are also shown as horizontal lines for reference.
More detailed per-task training dynamics are provided in Appendix~\ref{sec:training_dynamics_each_task}.

\noindent\textbf{Jagle is effective for Japanese tasks.}
The model trained on Jagle achieves an average score on Japanese tasks more than 20 points higher than the model trained on FineVision alone.
Compared to baseline models, the Jagle-trained model surpasses InternVL3.5-2B on the Japanese task average and comes within 5 points of Qwen3-VL-2B-Instruct.
At 60,000 training steps, the Jagle-trained model has consumed approximately 150B tokens in total, which is less than one-tenth of the roughly 2T tokens used by Qwen3-VL during multimodal post-training~\citep{bai2025qwen3vl}. Notably, performance continues to improve through the final training step with no sign of saturation, suggesting that further training could bring the model to a level comparable to or exceeding Qwen3-VL-2B-Instruct.
We also verify via pHash-based image matching that the overlap between Jagle training images and evaluation benchmark images is negligible ($<$0.002\%; Appendix~\ref{sec:contamination_details}).
These results demonstrate that Jagle is a practically useful dataset for improving Japanese performance.

\noindent\textbf{Impact on English tasks.}
When Jagle is combined with FineVision, the average score on English tasks is higher than that of the FineVision-only setting, demonstrating that Jagle not only avoids degrading but improves English task performance.
This result runs counter to the curse of multilinguality~\citep{shen2024curseofmultilinguality}. A similar finding was reported in FineVision~\citep{wiedmann2025finevision}, where adding Chinese data improved English performance, attributed to increased data diversity.
We hypothesize that incorporating Jagle similarly increases diversity and thereby has a positive effect on English task performance.
On the other hand, the Japanese task average is higher for Jagle alone than for Jagle combined with FineVision. The reason for this discrepancy between JA Avg and EN Avg is not entirely clear, though it may partly stem from the smaller data size of Jagle relative to FineVision; we leave a deeper investigation to future work.

\begin{table}[t]
\centering
\footnotesize
\setlength{\tabcolsep}{3pt}
\begin{tabular}{lllcccc}
\toprule
\textbf{Model} & \textbf{Base LLM} & \textbf{Training Data} & \textbf{Steps} & \textbf{JA Avg} & \textbf{EN Avg} & \textbf{Avg} \\
\midrule
\rowcolor{gray!15}\multicolumn{7}{l}{\textit{2B-scale models}} \\
\multirow{3}{*}{Jagle-VL-2.2B (ours)} & \multirow{3}{*}{Qwen3-1.7B} & FineVision & 60k & 29.6 {\scriptsize$\pm$0.0} & 57.7 & 43.7 {\scriptsize$\pm$0.0} \\
 & & Jagle & 60k & 51.3 {\scriptsize$\pm$0.1} & 52.3 & 51.8 {\scriptsize$\pm$0.0} \\
 & & Jagle+FineVision & 60k & 50.1 {\scriptsize$\pm$0.0} & 59.6 & 54.9 {\scriptsize$\pm$0.0} \\
\midrule
Qwen3-VL-2B-Instruct & Qwen3-1.7B & & & \textbf{56.0} {\scriptsize$\pm$0.2} & 67.0 & \textbf{61.5} {\scriptsize$\pm$0.1} \\
InternVL3.5-2B & Qwen3-1.7B & & & 47.0 {\scriptsize$\pm$0.2} & \textbf{68.0} & 57.5 {\scriptsize$\pm$0.1} \\
\midrule
\rowcolor{gray!15}\multicolumn{7}{l}{\textit{8B-scale and larger models}} \\
\multirow{2}{*}{LLM-jp-4-VL 9B beta (ours)} & \multirow{2}{*}{LLM-jp-4-8B} & Jagle+FineVision & 60k & 69.1 {\scriptsize$\pm$0.1} & 70.4 & 69.7 {\scriptsize$\pm$0.0} \\
 & & Jagle+FineVision & 90k & 70.8 {\scriptsize$\pm$0.1} & 71.1 & 71.0 {\scriptsize$\pm$0.1} \\
\midrule
Qwen3-VL-8B-Instruct & Qwen3-8B & & & \textbf{71.1} {\scriptsize$\pm$0.2} & \textbf{77.0} & \textbf{74.1} {\scriptsize$\pm$0.1} \\
InternVL3.5-8B & Qwen3-8B & & & 56.9 {\scriptsize$\pm$0.2} & 74.0 & 65.4 {\scriptsize$\pm$0.1} \\
LLM-jp-3 VILA 14B & LLM-jp-3-13B & & & 33.0 {\scriptsize$\pm$0.2} & 43.7 & 38.4 {\scriptsize$\pm$0.1} \\
\bottomrule
\end{tabular}
\caption{Comparison of models trained with Jagle against existing VLMs. We report the mean and standard deviation over three evaluation runs (omitted for EN Avg, where evaluation is deterministic). Bold indicates the best score in each column within each scale group. Per-task results of all models are provided in Appendix~\ref{sec:per_task_results}.}
\label{tab:model-comparison}
\end{table}

\noindent\textbf{Comparison with a model trained on a prior Japanese dataset.}
To compare Jagle with previously released Japanese multimodal post-training datasets, we additionally evaluate LLM-jp-3 VILA 14B~\citep{sasagawa-etal-2025-constructing}, a VLM trained on the LLM-jp-3 VILA dataset.
Table~\ref{tab:model-comparison} shows the results. Despite being a 14B-scale model, LLM-jp-3 VILA 14B underperforms Jagle-VL-2.2B trained on Jagle and on Jagle+FineVision across the Japanese, English, and overall averages.
A likely contributing factor is the substantial difference in dataset scale and diversity: the LLM-jp-3 VILA dataset contains approximately 0.4M examples, whereas Jagle contains 9.2M instances spanning substantially broader domains and task types.
Note, however, that this comparison is not perfectly controlled, as the training settings and model architectures differ.

\subsection{Generality across model backbones and scales}
\label{sec:generality_backbones}

To validate that the effectiveness of Jagle is not tied to the specific Qwen3-1.7B backbone, we additionally replace the base LLM with LLM-jp-4-8B-instruct~\citep{llmjp2024llmjp}, a Japanese-centric LLM trained on large-scale Japanese data.
The other architectural components remain the same as in Section~\ref{sec:training_setup}. We train the model on the Jagle + FineVision mixture and extend the training from 60k to 90k steps to accommodate the larger model scale. We refer to the final model trained for 90k steps as LLM-jp-4-VL 9B beta, and additionally report its intermediate checkpoint at 60k steps.
We compare it with Qwen3-VL-8B-Instruct~\citep{bai2025qwen3vl} and InternVL3.5-8B~\citep{wang2025internvl35}, strong VLMs of similar size.

The results are shown in Table~\ref{tab:model-comparison}. The model trained for 90k steps achieves a Japanese average score of 70.8, outperforming InternVL3.5-8B (56.9) and reaching performance comparable to Qwen3-VL-8B-Instruct (71.1).
These results demonstrate that the effectiveness of Jagle is not tied to a specific backbone or scale.
Moreover, performance at 90k steps is consistently higher than at 60k steps across the Japanese, English, and overall averages, indicating that the model has not yet saturated at 60k steps and that further training remains beneficial.

\section{Conclusion}

In this work, we introduced \method{}, the largest Japanese multimodal post-training dataset, comprising 9.2M instances, constructed with a scalable pipeline that collects heterogeneous data sources and generates diverse QA data through multiple strategies.
Experiments demonstrated that Jagle effectively improves Japanese task performance without degrading English performance when combined with FineVision, and that these benefits generalize across model backbones and scales.
We hope this work contributes to the advancement of multilingual VLM research.

\section*{Limitations}

\noindent\textbf{Optimal Dataset Mixture.}
Jagle is constructed from diverse data sources to cover all five categories; however, we do not explicitly control or optimize the proportion of each category. Prior works suggest that dataset mixture plays a crucial role in model performance~\citep{tong2024cambrian,chen2026olmix}. Exploring optimal mixtures across categories is an important direction for future work.

\noindent\textbf{Data Filtering.}
We employ Qwen3-VL to generate question-answer pairs, but model-based generation methods are known to suffer from issues such as hallucination and limited diversity~\citep{niklaus2026finephrase}. Addressing these challenges could further improve dataset quality. For large-scale datasets, both rule-based and model-based filtering strategies are promising approaches~\citep{nvidia2025nvidianemotron3efficient}.

\noindent\textbf{Omitted Categories.}
In this work, we exclude several categories such as Grounding \& Counting, Math, Science, and Text-only tasks. To develop general-purpose VLMs, future work should incorporate these categories.

\section*{Acknowledgements}
In this research work, we used the ``mdx: a platform for building data-empowered society''.
We used ABCI 3.0 provided by AIST and AIST Solutions with support from ``ABCI 3.0 Development Acceleration Use''.
We used a list of website URLs provided by the National Diet Library, which had been collected through its Web Archiving Project (WARP).
This research was supported by the Science Tokyo Support Program for Doctoral Students,
funded by the Universities for International Research Excellence.

\bibliography{colm2026_conference}
\bibliographystyle{colm2026_conference}

\appendix

\section{Detailed Statistics of Jagle Dataset Subsets}
\label{sec:detailed_statistic_jagle}
Table~\ref{tab:dataset-stats} provides detailed statistics for each subset of the Jagle dataset.

\begin{table}[t]
\centering
\setlength{\tabcolsep}{2pt}
\caption{Detailed statistics of each subset in \method{}, including the number of samples, unique images, turns, and answer tokens.}
\label{tab:dataset-stats}
\small
\begin{tabular}{p{2cm}lrrrr}
\toprule
\textbf{Category} & \textbf{Subset Name} & \textbf{Samples} & \textbf{Unique Images} & \textbf{Turns} & \textbf{Answer Tokens} \\
\midrule
\multirow{6}{*}{General VQA} & japanese-photos-VQA & 1,163 & 1,163 & 5,775 & 227,515 \\
 & JSQuAD-Vision-filterd-15k & 14,790 & 14,790 & 39,319 & 833,581 \\
 & ja-vg-vqa & 99,202 & 99,199 & 793,616 & 15,247,119 \\
 & llava-instruct-ja-qwen3vl & 155,657 & 80,958 & 386,229 & 60,802,343 \\
 & WIKI-JA-VQA & 936,871 & 936,871 & 4,679,651 & 178,987,472 \\
 & WAON-VQA & 1,912,226 & 1,912,213 & 9,541,389 & 302,272,463 \\
\midrule
\multirow{3}{*}{Chart \& Table}
 & tat-qa-ja-translated-2k & 2,180 & 2,180 & 13,089 & 444,245 \\
& WAON-Chart-VQA & 98,791 & 98,703 & 481,076 & 27,205,736 \\
 & plotqa-ja-translated-153k & 152,912 & 152,911 & 19,597,685 & 401,888,887 \\
\midrule
Captioning  & WIKI-JA-Captioning & 993,006 & 993,006 & 1,986,012 & 198,679,655 \\
\& Knowledge & NDL-PDF-detail & 1,000,000 & 986,031 & 1,000,000 & 1,440,399,034 \\
\midrule
\multirow{3}{*}{OCR QA} & WAON-OCR-VQA & 871,847 & 871,617 & 3,399,288 & 72,444,578 \\
 & NDL-PDF-simple & 1,000,000 & 987,042 & 8,824,404 & 311,860,431 \\
  & FinePDFs-Edu-JA-VQA & 1,666,699 & 1,666,424 & 7,910,259 & 341,396,954 \\
\midrule
\multirow{2}{*}{Naive OCR}
 & JSSODa-train-18k-v2 & 17,991 & 17,991 & 17,991 & 8,472,460 \\
 & e-Gov-OCR & 31,190 & 30,134 & 31,190 & 32,292,822 \\
 & Nemotron-Wiki-Ja-OCR & 199,999 & 199,999 &  199,999 & 117,455,487\\
\midrule
Total & & 9,154,524 & 9,051,232 & 58,906,972 & 3,510,910,782 \\
\bottomrule
\end{tabular}
\end{table}

\section{Image Embedding Analysis}
\label{sec:image_embedding_analysis}

To analyze the visual diversity of images in Jagle, we randomly sampled 5,000 images from the dataset and computed their image embeddings, which were subsequently visualized using t-SNE. The embeddings were extracted using SigLIP2-so400M-Patch16-512~\citep{tschannen2025siglip2} and projected into two dimensions via t-SNE~\citep{maaten2008t-sne}.

\begin{figure}[t]
\begin{center}
\includegraphics[width=0.8\linewidth]{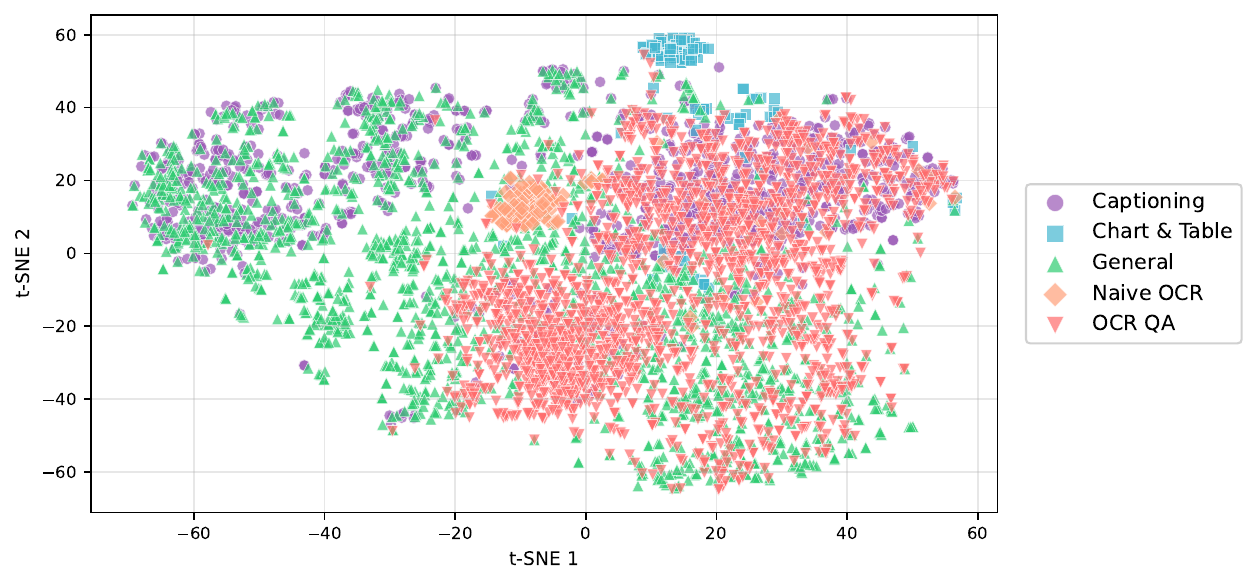}
\caption{t-SNE visualization of SigLIP2 image embeddings for 5,000 images randomly sampled from \method{}. Chart \& Table and Naive OCR images form distinct clusters, while General VQA, Captioning, and OCR QA images are largely intermingled.}
\label{fig:jagel-clustering}
\end{center}
\end{figure}

The results are shown in Figure~\ref{fig:jagel-clustering}. Samples from the General VQA, Captioning, and OCR QA categories are largely intermingled, whereas Chart \& Table and Naive OCR images form relatively compact clusters. This pattern likely arises because Chart \& Table and Naive OCR images, which primarily consist of charts, tables, and document-style content, exhibit visual characteristics that differ substantially from the natural images that dominate categories such as General VQA and Captioning \& Knowledge.

\section{Licenses of Source Datasets and Models}
\label{sec:licenses}

We construct Jagle with careful attention to licensing compatibility so that the dataset can be released and used in commercially permissible settings as much as possible.
All source datasets are distributed under licenses that permit commercial use or are legally usable for information analysis purposes under Japanese copyright law, and all models used for QA generation and preprocessing are distributed under commercially permissive licenses (e.g., Apache 2.0).
Tables~\ref{tab:licenses_datasets} and~\ref{tab:licenses_models} list the licenses of the source datasets used to construct Jagle and the models used for QA generation and preprocessing, respectively.

\begin{table}[t]
\centering
\small
\begin{tabular}{lp{7cm}}
\toprule
\textbf{Source Dataset} & \textbf{License} \\
\midrule
WAON~{\scriptsize\citep{sugiura2025waon}} & Apache 2.0 (use limited to information analysis under Article 30-4 of the Japanese Copyright Act) \\
japanese-photos~{\scriptsize\citep{ThePioneer2024japanese-photos}} & CC0 1.0 \\
ja-vg-vqa~{\scriptsize\citep{shimizu-etal-2018-visual}} & CC BY 4.0 \\
llava-instruct~{\scriptsize\citep{liu2023llava}} & CC BY 4.0 \\
JSQuAD~{\scriptsize\citep{kurihara-etal-2022-jglue}} & CC BY-SA 4.0 \\
Wiki-JA-Pair (ours) & CC BY-SA 4.0 (derived from Japanese Wikipedia) \\
FinePDFs-Edu~{\scriptsize\citep{kydlicek2025finepdfs}} & ODC-BY \\
PlotQA~{\scriptsize\citep{Methani_2020_plotqa}} & CC BY 4.0 \\
TAT-QA~{\scriptsize\citep{zhu-etal-2021-tatqa}} & CC BY 4.0 \\
NDL WARP PDF~{\scriptsize\citep{ndl-warp-pdf}} & Use limited to information analysis under Article 30-4 of the Japanese Copyright Act \\
e-Gov~{\scriptsize\citep{egovjp}} & Government of Japan Standard Terms of Use 2.0 \\
JSSODa~{\scriptsize\citep{sasagawa-etal-2026-evaluating}} & CC BY 4.0 \\
Nemotron-VLM-Dataset-v2~{\scriptsize\citep{nvidia2025nvidianemotronnanov2}} & CC BY 4.0 \\
\bottomrule
\end{tabular}
\caption{Licenses of the source datasets used to construct \method{}.}
\label{tab:licenses_datasets}
\end{table}

\begin{table}[t]
\centering
\small
\begin{tabular}{ll}
\toprule
\textbf{Model} & \textbf{License} \\
\midrule
Qwen3-VL-235B-A22B-Instruct~{\scriptsize\citep{bai2025qwen3vl}} & Apache 2.0 \\
LLM-jp-3-13B-instruct~{\scriptsize\citep{llmjp2024llmjp}} & Apache 2.0 \\
PaddleOCR-VL~{\scriptsize\citep{cui2025paddleocrvl}} & Apache 2.0 \\
\bottomrule
\end{tabular}
\caption{Licenses of the models used for QA generation and preprocessing.}
\label{tab:licenses_models}
\end{table}

\section{Benchmark Contamination Analysis}
\label{sec:contamination_details}

Since Jagle is constructed from large-scale web-based sources such as Japanese Wikipedia and Common Crawl, its training images may overlap with those used in evaluation benchmarks.
To investigate potential benchmark contamination, we perform pHash-based image matching between Jagle training images and the images of the Japanese evaluation benchmarks, counting the number of unique benchmark images with matching perceptual hashes for each subset.
Table~\ref{tab:contamination} shows, for each subset of Jagle, the number of unique benchmark images with matching perceptual hashes and the breakdown of the matching benchmarks.

Only 4 of the 17 subsets contain images matching evaluation benchmarks, amounting to 153 unique benchmark images in total, which accounts for less than 0.002\% of the entire Jagle dataset.
Importantly, the corresponding QA pairs are not crawled from the web but independently generated by our pipeline; therefore, even for overlapping images, the benchmark questions and answers themselves are not present in the training data.
Moreover, Jagle consistently improves performance on benchmarks without any detected overlap (e.g., CC-OCR-JA, JA-VLM-Bench-In-the-Wild, and BusinessSlideVQA), suggesting that the observed gains cannot be explained by benchmark overlap.

Manual inspection of the matched images reveals that they predominantly consist of historical portraits (e.g., samurai figures) and traditional Japanese paintings---culturally significant images that appear broadly across Japanese Wikipedia and are therefore likely to co-occur in both training and evaluation sources.
Since these images are not replaceable with alternatives, and since their QA annotations were independently generated by our pipeline rather than sourced from benchmarks, we regard these overlaps as largely unavoidable and unlikely to materially affect our assessment of Jagle's effectiveness.

\begin{table}[t]
\centering
\setlength{\tabcolsep}{3pt}
\resizebox{\linewidth}{!}{%
\begin{tabular}{lrl}
\toprule
\textbf{Subset Name} & \textbf{Unique pHash Matches} & \textbf{Matching Benchmark Breakdown} \\
\midrule
japanese-photos-VQA & 0 & -- \\
JSQuAD-Vision-filterd-15k & 0 & -- \\
ja-vg-vqa & 0 & -- \\
llava-instruct-ja-qwen3vl & 0 & -- \\
WIKI-JA-VQA & 141 & JMMMU: 129, CVQA-JA: 10, Heron-Bench: 2 \\
WAON-VQA & 5 & JMMMU: 5 \\
tat-qa-ja-translated-2k & 0 & -- \\
WAON-Chart-VQA & 0 & -- \\
plotqa-ja-translated-153k & 0 & -- \\
WIKI-JA-Captioning & 149 & JMMMU: 135, CVQA-JA: 11, Heron-Bench: 3 \\
NDL-PDF-detail & 0 & -- \\
WAON-OCR-VQA & 0 & -- \\
NDL-PDF-simple & 0 & -- \\
FinePDFs-Edu-JA-VQA & 1 & JDocQA: 1 \\
JSSODa-train-18k-v2 & 0 & -- \\
e-Gov-OCR & 0 & -- \\
Nemotron-Wiki-Ja-OCR & 0 & -- \\
\midrule
Overall & 153 & JMMMU: 136, CVQA-JA: 11, JDocQA: 3, Heron-Bench: 3 \\
\bottomrule
\end{tabular}%
}
\caption{Per-subset results of the pHash-based contamination analysis between Jagle training images and Japanese evaluation benchmark images. For each subset, we report the number of unique benchmark images with matching perceptual hashes. Since the same benchmark image can match images in multiple subsets, the Overall row counts unique benchmark images across all subsets.}
\label{tab:contamination}
\end{table}

\section{Per-Task Evaluation Results}
\label{sec:per_task_results}

Tables~\ref{tab:per_task_ja} and~\ref{tab:per_task_en} report per-task results of all evaluated models on the 10 Japanese and 10 English benchmarks, respectively.
We report the mean and standard deviation over three evaluation runs; standard deviations are omitted for tasks with deterministic evaluation.
The observed improvements of Jagle-trained models over InternVL3.5-2B are substantially larger than the run-to-run variance, suggesting that the gains are robust rather than artifacts of evaluation noise.

\begin{table}[t]
\centering
\setlength{\tabcolsep}{4pt}
\resizebox{\linewidth}{!}{%
\begin{tabular}{lccccc}
\toprule
\textbf{Model} & \textbf{CC-OCR-JA} & \textbf{JA-VLM-Bench} & \textbf{Heron-Bench} & \textbf{CVQA-JA} & \textbf{MECHA-JA} \\
\midrule
\rowcolor{gray!15}\multicolumn{6}{l}{\textit{2B-scale models}} \\
Jagle-VL-2.2B (FineVision) & 18.2 {\scriptsize$\pm$0.4} & 34.7 {\scriptsize$\pm$0.0} & 31.8 {\scriptsize$\pm$0.0} & 45.0 & 44.1 \\
Jagle-VL-2.2B (Jagle) & \textbf{55.2} {\scriptsize$\pm$0.0} & \textbf{63.3} {\scriptsize$\pm$0.0} & 53.4 {\scriptsize$\pm$0.0} & 46.0 & 43.6 \\
Jagle-VL-2.2B (Jagle+FineVision) & 49.7 {\scriptsize$\pm$0.0} & 61.2 {\scriptsize$\pm$0.0} & 51.1 {\scriptsize$\pm$0.0} & 51.0 & 46.1 \\
\midrule
Qwen3-VL-2B-Instruct & 54.3 {\scriptsize$\pm$1.1} & 55.1 {\scriptsize$\pm$0.0} & \textbf{54.5} {\scriptsize$\pm$0.0} & \textbf{51.5} & \textbf{48.5} \\
InternVL3.5-2B & 52.2 {\scriptsize$\pm$0.4} & 43.5 {\scriptsize$\pm$1.2} & 41.3 {\scriptsize$\pm$0.7} & 40.0 & 44.3 \\
\midrule
\rowcolor{gray!15}\multicolumn{6}{l}{\textit{8B-scale and larger models}} \\
LLM-jp-4-VL 9B beta (60k) & 63.0 {\scriptsize$\pm$0.4} & \textbf{69.4} {\scriptsize$\pm$0.0} & \textbf{66.7} {\scriptsize$\pm$0.7} & 64.0 & 70.8 \\
LLM-jp-4-VL 9B beta (90k) & 68.0 {\scriptsize$\pm$0.4} & \textbf{69.4} {\scriptsize$\pm$0.0} & 63.6 {\scriptsize$\pm$0.0} & \textbf{68.0} & \textbf{72.0} \\
\midrule
Qwen3-VL-8B-Instruct & \textbf{74.3} {\scriptsize$\pm$0.4} & 66.0 {\scriptsize$\pm$1.2} & 61.7 {\scriptsize$\pm$1.7} & 62.0 & 63.9 \\
InternVL3.5-8B & 51.0 {\scriptsize$\pm$0.7} & 46.9 {\scriptsize$\pm$0.0} & 51.1 {\scriptsize$\pm$0.0} & 49.0 & 57.2 \\
LLM-jp-3 VILA 14B & 14.9 {\scriptsize$\pm$0.4} & 52.4 {\scriptsize$\pm$1.2} & 50.8 {\scriptsize$\pm$0.7} & 49.5 & 48.1 \\
\bottomrule
\end{tabular}%
}

\vspace{2mm}

\resizebox{\linewidth}{!}{%
\begin{tabular}{lcccccc}
\toprule
\textbf{Model} & \textbf{JA-Multi-Image} & \textbf{JMMMU} & \textbf{JDocQA} & \textbf{BusinessSlideVQA} & \textbf{JGraphQA} & \textbf{JA Avg} \\
\midrule
\rowcolor{gray!15}\multicolumn{7}{l}{\textit{2B-scale models}} \\
Jagle-VL-2.2B (FineVision) & 28.3 {\scriptsize$\pm$0.0} & 35.7 & 15.5 {\scriptsize$\pm$0.1} & 14.4 {\scriptsize$\pm$0.0} & 28.4 {\scriptsize$\pm$0.3} & 29.6 {\scriptsize$\pm$0.0} \\
Jagle-VL-2.2B (Jagle) & 39.6 {\scriptsize$\pm$0.0} & 37.7 & 60.8 {\scriptsize$\pm$0.2} & 49.5 {\scriptsize$\pm$0.5} & 64.3 {\scriptsize$\pm$0.0} & 51.3 {\scriptsize$\pm$0.1} \\
Jagle-VL-2.2B (Jagle+FineVision) & 39.6 {\scriptsize$\pm$0.0} & \textbf{39.1} & 54.4 {\scriptsize$\pm$0.1} & 43.4 {\scriptsize$\pm$0.3} & 65.5 {\scriptsize$\pm$0.3} & 50.1 {\scriptsize$\pm$0.0} \\
\midrule
Qwen3-VL-2B-Instruct & \textbf{63.5} {\scriptsize$\pm$1.1} & 35.0 & \textbf{68.5} {\scriptsize$\pm$0.3} & \textbf{50.2} {\scriptsize$\pm$0.3} & \textbf{79.1} {\scriptsize$\pm$0.0} & \textbf{56.0} {\scriptsize$\pm$0.2} \\
InternVL3.5-2B & 39.6 {\scriptsize$\pm$0.0} & 38.6 & 52.7 {\scriptsize$\pm$0.1} & 43.7 {\scriptsize$\pm$1.0} & 74.0 {\scriptsize$\pm$0.0} & 47.0 {\scriptsize$\pm$0.2} \\
\midrule
\rowcolor{gray!15}\multicolumn{7}{l}{\textit{8B-scale and larger models}} \\
LLM-jp-4-VL 9B beta (60k) & 86.8 {\scriptsize$\pm$0.0} & 52.9 & 68.4 {\scriptsize$\pm$0.1} & 61.9 {\scriptsize$\pm$0.3} & 86.7 {\scriptsize$\pm$0.0} & 69.1 {\scriptsize$\pm$0.1} \\
LLM-jp-4-VL 9B beta (90k) & \textbf{88.7} {\scriptsize$\pm$0.0} & \textbf{54.0} & 73.7 {\scriptsize$\pm$0.1} & 61.4 {\scriptsize$\pm$1.2} & \textbf{89.1} {\scriptsize$\pm$0.3} & 70.8 {\scriptsize$\pm$0.1} \\
\midrule
Qwen3-VL-8B-Instruct & 86.8 {\scriptsize$\pm$0.0} & 51.7 & \textbf{86.5} {\scriptsize$\pm$0.1} & \textbf{70.8} {\scriptsize$\pm$0.0} & 87.2 {\scriptsize$\pm$0.5} & \textbf{71.1} {\scriptsize$\pm$0.2} \\
InternVL3.5-8B & 61.6 {\scriptsize$\pm$1.1} & 49.8 & 59.0 {\scriptsize$\pm$0.1} & 58.0 {\scriptsize$\pm$0.7} & 85.2 {\scriptsize$\pm$0.5} & 56.9 {\scriptsize$\pm$0.2} \\
LLM-jp-3 VILA 14B & 50.9 {\scriptsize$\pm$0.0} & 28.6 & 8.6 {\scriptsize$\pm$0.1} & 9.1 {\scriptsize$\pm$0.5} & 17.3 {\scriptsize$\pm$0.0} & 33.0 {\scriptsize$\pm$0.2} \\
\bottomrule
\end{tabular}%
}
\caption{Per-task results of all evaluated models on the 10 Japanese benchmarks (split into two column blocks). We report the mean and standard deviation over three evaluation runs; standard deviations are omitted for tasks with deterministic evaluation. Bold indicates the best score in each column within each scale group; within each group, our models and existing VLMs are separated by a horizontal line.}
\label{tab:per_task_ja}
\end{table}

\begin{table}[t]
\centering
\setlength{\tabcolsep}{4pt}
\resizebox{\linewidth}{!}{%
\begin{tabular}{lccccc}
\toprule
\textbf{Model} & \textbf{TextVQA} & \textbf{OK-VQA} & \textbf{MMMU} & \textbf{ScienceQA} & \textbf{DocVQA} \\
\midrule
\rowcolor{gray!15}\multicolumn{6}{l}{\textit{2B-scale models}} \\
Jagle-VL-2.2B (FineVision) & 69.4 & 32.7 & 44.2 & 83.4 & 79.2 \\
Jagle-VL-2.2B (Jagle) & 66.6 & 31.8 & 44.7 & 71.0 & 78.7 \\
Jagle-VL-2.2B (Jagle+FineVision) & 72.4 & 35.0 & 43.0 & 83.6 & 83.8 \\
\midrule
Qwen3-VL-2B-Instruct & \textbf{80.6} & \textbf{52.9} & 43.9 & 80.5 & \textbf{92.7} \\
InternVL3.5-2B & 75.3 & 50.5 & \textbf{51.5} & \textbf{91.7} & 88.9 \\
\midrule
\rowcolor{gray!15}\multicolumn{6}{l}{\textit{8B-scale and larger models}} \\
LLM-jp-4-VL 9B beta (60k) & 80.6 & 50.5 & 46.3 & 93.4 & 91.4 \\
LLM-jp-4-VL 9B beta (90k) & 81.8 & 50.1 & 46.8 & 94.1 & 92.7 \\
\midrule
Qwen3-VL-8B-Instruct & \textbf{84.0} & \textbf{61.1} & 57.7 & 94.4 & \textbf{95.7} \\
InternVL3.5-8B & 77.7 & 59.0 & \textbf{60.6} & \textbf{97.4} & 91.6 \\
LLM-jp-3 VILA 14B & 60.9 & 50.5 & 31.4 & 65.0 & 40.8 \\
\bottomrule
\end{tabular}%
}

\vspace{2mm}

\resizebox{\linewidth}{!}{%
\begin{tabular}{lcccccc}
\toprule
\textbf{Model} & \textbf{InfoVQA} & \textbf{AI2D} & \textbf{ChartQA} & \textbf{BLINK} & \textbf{RealWorldQA} & \textbf{EN Avg} \\
\midrule
\rowcolor{gray!15}\multicolumn{7}{l}{\textit{2B-scale models}} \\
Jagle-VL-2.2B (FineVision) & 52.5 & 65.3 & 54.2 & 41.0 & 55.6 & 57.7 \\
Jagle-VL-2.2B (Jagle) & 43.9 & 64.0 & 34.4 & 40.0 & 47.5 & 52.3 \\
Jagle-VL-2.2B (Jagle+FineVision) & 55.9 & 66.2 & 57.7 & 41.5 & 57.4 & 59.6 \\
\midrule
Qwen3-VL-2B-Instruct & \textbf{71.9} & 76.3 & 66.5 & 41.3 & \textbf{63.9} & 67.0 \\
InternVL3.5-2B & 67.1 & \textbf{78.6} & \textbf{67.4} & \textbf{47.6} & 61.2 & \textbf{68.0} \\
\midrule
\rowcolor{gray!15}\multicolumn{7}{l}{\textit{8B-scale and larger models}} \\
LLM-jp-4-VL 9B beta (60k) & 71.9 & 75.3 & 75.0 & 50.3 & 69.3 & 70.4 \\
LLM-jp-4-VL 9B beta (90k) & 74.4 & 75.6 & 75.9 & 48.6 & \textbf{71.2} & 71.1 \\
\midrule
Qwen3-VL-8B-Instruct & \textbf{82.2} & \textbf{85.1} & 75.0 & \textbf{65.5} & 69.4 & \textbf{77.0} \\
InternVL3.5-8B & 74.0 & 84.2 & \textbf{78.1} & 53.1 & 64.3 & 74.0 \\
LLM-jp-3 VILA 14B & 27.4 & 46.2 & 26.8 & 40.2 & 48.1 & 43.7 \\
\bottomrule
\end{tabular}%
}
\caption{Per-task results of all evaluated models on the 10 English benchmarks (split into two column blocks), where evaluation is deterministic. Bold indicates the best score in each column within each scale group; within each group, our models and existing VLMs are separated by a horizontal line.}
\label{tab:per_task_en}
\end{table}

\section{Prompts for QA Generation}
\label{sec:prompt_for_qa_generation}

Below we present the prompts used in the VLM-based QA generation method for each dataset.

\begin{Prompt}{QA Generation Prompt (WIKI-JA-Captioning)}
あなたは画像理解に優れたAIアシスタントです.
以下の情報が与えられます:

- 画像(Wikipedia記事に掲載されたもの)

- 画像のキャプション

- 記事のタイトル

- 記事の本文

これらの情報をヒントとして利用しながら, **画像の内容を説明させるための質問と答え(QAペア)**を2組作成してください. 1組は「簡潔な説明」を引き出す質問, もう1組は「詳細な説明」を引き出す質問にしてください.

条件：

1. キャプションやタイトル、記事本文を与えられていない第三者が画像および一般知識だけで答えられる内容にしてください. ただし, キャプションおよびタイトルの情報は直接含めても構いません.

- 例：「この画像には何が写っていますか？簡潔に答えてください。」,「この画像に写っているものを詳しく説明してください。」

2. 質問と答えは自然な日本語にしてください.

3. 推測や意見を含めないでください.

出力形式はJSON構造:

\{\{
"conversations": [
        \{\{"from": "human", "value": "質問(簡潔説明用)"\}\},
        \{\{"from": "gpt", "value": "答え(簡潔説明)"\}\}
        \{\{"from": "human", "value": "質問(詳細説明用)"\}\},
        \{\{"from": "gpt", "value": "答え(詳細説明)"\}\},
]
\}\}

入力情報:

- 画像のキャプション：{caption}

- 記事タイトル：{title}

- 記事本文：{text}

出力:
\end{Prompt}

\begin{Prompt}{QA Generation Prompt (WAON-Chart-VQA)}
あなたのタスクは、与えられた画像に基づいて、適切な質問とその答えを生成することです.

以下の情報が与えられます:

- 画像

もし画像に棒グラフや折れ線グラフ、表、ダイアグラムなどの図表が含まれていない場合や画質が悪すぎる画像、日本語ではない画像については、空のjsonを返してください。
画像が図表画像であれば、**画像の内容と一般知識で正しく答えられる質問と答え(QAペア)**を3〜5組作成してください.

質問の条件:

1. 第三者が画像および一般知識だけで答えられる内容にしてください.

2. 以下のタイプの質問をバランスよく含めてください：

   - 図表が何を表しているかに関する質問

   - 図表中の数値やテキストを抜き出す質問

   - 色や形、構造、関係などが必要な質問

   - 図表のデータを解釈・分析する質問

3. 質問と答えは自然な日本語にしてください.

4. 推測や意見を含めないでください.

出力形式はJSON構造:

\{\{
"conversations": [
    \{\{"from":"human","value":"質問1"\}\},
    \{\{"from":"gpt","value":"答え1"\}\},
    ...
]
\}\}

出力:
\end{Prompt}

\begin{Prompt}{QA Generation Prompt (Finepdfs-edu-ja-vqa)}
あなたのタスクは、与えられた画像に基づいて、適切な質問とその答えを生成することです.

以下の情報が与えられます:

- 画像

もし日本語ではない画像や画質が悪い画像の場合、空のjsonを返してください。
そうでなければ、**画像の内容と一般知識で正しく答えられる質問と答え(QAペア)**を3〜5組作成してください.

質問の条件:

1. 第三者が画像および一般知識だけで答えられる内容にしてください.

2. 以下のタイプの質問をバランスよく含めてください：

   - ドキュメントの内容を読み取る質問

   - ドキュメント中の図表や画像に関する質問

   - テキストや数値を読み取る質問

3. 質問と答えは自然な日本語にしてください.

4. 推測や意見を含めないでください.

出力形式はJSON構造:

\{\{
"conversations": [
    \{\{"from":"human","value":"質問1"\}\},
    \{\{"from":"gpt","value":"答え1"\}\},
    ...
]
\}\}

出力:
\end{Prompt}

\begin{Prompt}{QA Generation Prompt (japanese-photos-VQA)}
あなたは画像理解に優れたAIアシスタントです.

以下の情報が与えられます:

- 画像

これらの情報をヒントとして利用しながら, **画像の内容と一般知識で正しく答えられる質問と答え(QAペア)**を3〜5組作成してください.

質問の条件:

1. 第三者が画像および一般知識だけで答えられる内容にしてください.

2. 以下のタイプの質問をバランスよく含めてください：

   - 画像の主題を問う

   - 画像中の物体・人物・建造物などを問う

   - 色・形・構造・数・位置・関係などの特徴を問う

   - 文字や記号が含まれていれば, それを読み取る質問

3. 質問と答えは自然な日本語にしてください.

4. 推測や意見を含めないでください.

出力形式はJSON構造:

\{\{
"conversations": [
    \{\{"from":"human","value":"質問1"\}\},
    \{\{"from":"gpt","value":"答え1"\}\},
    ...
]
\}\}

出力:
\end{Prompt}

\section{Training Dynamics on Each Task}
\label{sec:training_dynamics_each_task}

\begin{figure}[t]
\begin{center}
\includegraphics[width=\linewidth]{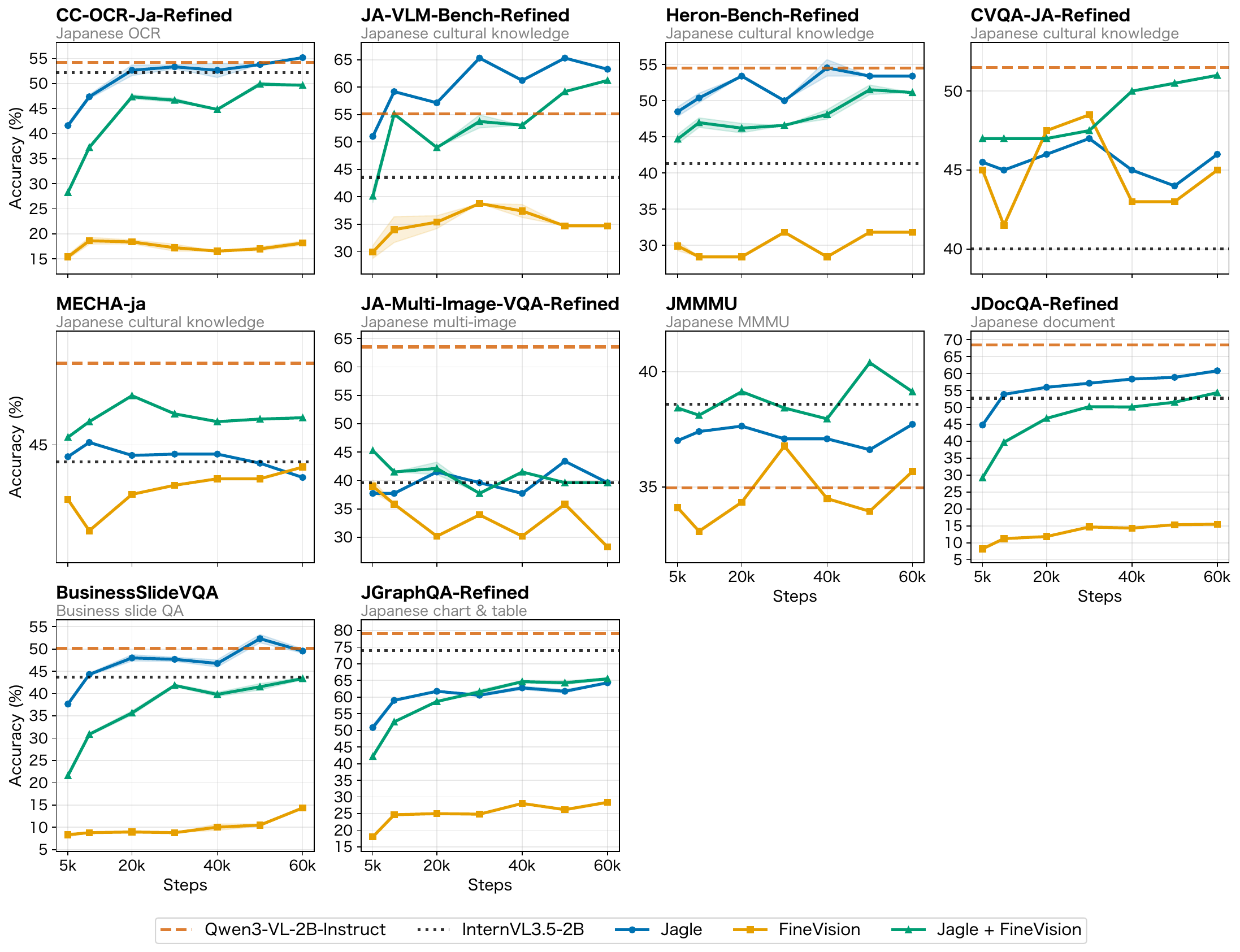}
\caption{Training dynamics under each data setting on each of the 10 Japanese benchmarks.}
\label{fig:training_curve_ja}
\end{center}
\end{figure}

\begin{figure}[t]
\begin{center}
\includegraphics[width=\linewidth]{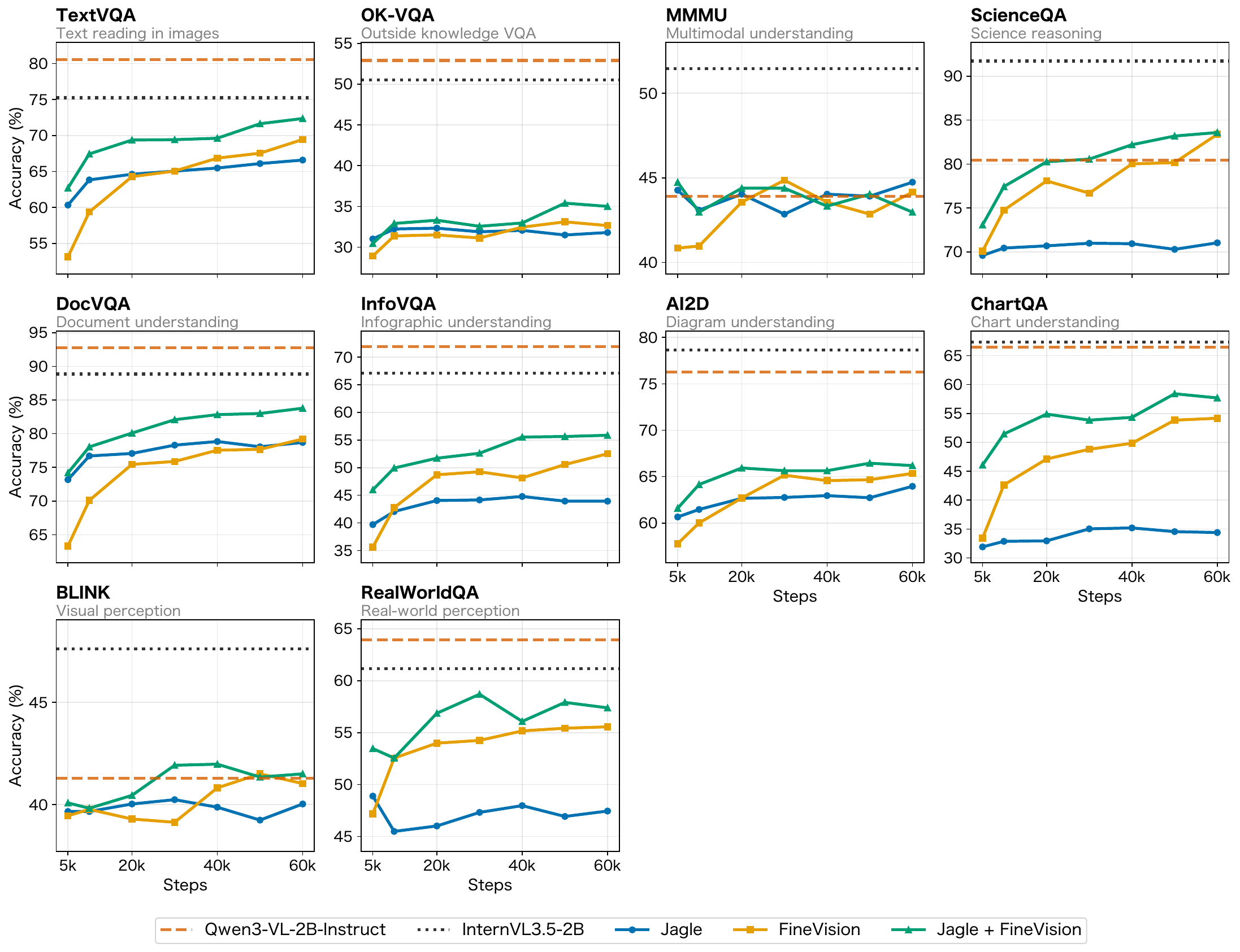}
\caption{Training dynamics under each data setting on each of the 10 English benchmarks.}
\label{fig:training_curve_en}
\end{center}
\end{figure}

Figures~\ref{fig:training_curve_ja} and~\ref{fig:training_curve_en} show the training dynamics for each of the 10 Japanese and 10 English tasks under the three data settings (Jagle, FineVision, and Jagle + FineVision), respectively.

\end{document}